\documentclass[runningheads]{llncs}
\usepackage{graphicx}
\usepackage{comment}
\usepackage{amsmath,amssymb} 
\usepackage{color}

\usepackage{times}
\usepackage{epsfig}
\usepackage{graphicx}
\usepackage{pifont}
\usepackage{algorithm}
\usepackage{algorithmic}
\usepackage{float}
\usepackage{xcolor}
\usepackage{stfloats}
\usepackage{subfigure}
\usepackage{multirow}
\usepackage{epstopdf}
\usepackage{setspace}
\usepackage{booktabs}
\usepackage{soul}
\usepackage{array}
\usepackage{caption}
\usepackage[utf8]{inputenc}

\usepackage[breaklinks=true,bookmarks=false]{hyperref}

\newcommand{\SE}{\mathbb{SE}}

\newcommand{\RR}{\mathbb{R}}
\newcommand{\LL}{\mathcal{L}}

\def\ie{\emph{i.e.}}

\begin{document}
\pagestyle{headings}
\mainmatter
\def\ECCVSubNumber{3301}  

\title{Feature-metric Loss for Self-supervised Learning of Depth and Egomotion} 


\titlerunning{Feature-metric Loss for Self-supervised Learning of Depth and Egomotion}
%
\author{
Chang Shu\inst{1}\thanks{This work is done when Chang Shu is an intern at DeepMotion.} \and
Kun Yu\inst{2} \and
Zhixiang Duan\inst{2} \and
Kuiyuan Yang\inst{2}
}
\authorrunning{C. Shu, K. Yu, Z. Duan and K. Yang}
%
\institute{Meituan Dianping Group\\
\and
DeepMotion\\
\email{shuchang02@meituan.com}\\
\email{\{kunyu,zhixiangduan,kuiyuanyang\}@deepmotion.ai}}
\maketitle

\begin{abstract}
Photometric loss is widely used for self-supervised depth and egomotion estimation. However, 
the loss landscapes induced by photometric differences are often problematic for optimization, caused by plateau landscapes for pixels in textureless regions or multiple local minima for less discriminative pixels. 
In this work, feature-metric loss is proposed and defined on feature representation, where the feature representation is also learned in a self-supervised manner and regularized by both first-order and  second-order derivatives to constrain the loss landscapes to form proper convergence basins.
Comprehensive experiments and detailed analysis via visualization demonstrate the effectiveness of the proposed feature-metric loss.
In particular, our method improves state-of-the-art methods on KITTI from 0.885 to 0.925 measured by $\delta_1$ for depth estimation, and significantly outperforms previous method for visual odometry.

\end{abstract}
\section{Introduction}
Estimating depth and egomotion from monocular camera is a fundamental and valuable task in computer vision, which has wide applications in augmented reality~\cite{dtam}, robotics navigation~\cite{desouza2002vision} and autonomous driving~\cite{menze2015object}. 
Though monocular camera is cheap and lightweight, the task is hard for conventional SfM/SLAM algorithms~\cite{dso,mur2015orb,pire2017s} and continues challenging deep learning based approaches~\cite{struct2depth,undeepvo,vomonodepth,sc-sfmlearner,sfmlearner}.

Deep learning for depth and egomotion estimation can be broadly categorized into supervised and self-supervised learning. 
For depth estimation, supervised learning takes images paired with depth maps as input~\cite{Eigen,dorn,bts}, where depth maps are sparsely collected from expensive LiDAR sensors ~\cite{kitti} or densely rendered from simulation engines~\cite{mayer2016large}, while supervision from LiDAR limits the generalization to new cameras and supervision from rendering limits the generalization to real scenes. 
For egomotion estimation, supervised signals come from trajectories computed by classical methods with high precision sensors like IMU and GPS, which are also costly and cannot guarantee absolute accuracy. 
Self-supervised learning unifies these two tasks into one framework, and only uses monocular videos as inputs, and supervision is from view synthesis~\cite{sfmlearner,GeoNet,vid2deep,monodepth,monodepth2}. 
The setup is simpler, and easy to generalize among cameras.

However, self-supervised approaches are still inferior to supervised ones by large margins when compared on standard benchmarks. 
The main problem lies in the weak supervision added as photometric loss, which is defined as the photometric difference between a pixel warped from source view by estimated depth and pose and the pixel captured in the target view. 
Nevertheless, small photometric loss does not necessarily guarantee accurate depth and pose, especially for pixels in textureless regions. 
The problem can be partially solved by adding smoothness loss on depth map, which encourages first-order smoothness~\cite{struct2depth,monodepth,monodepth2} or second-order smoothness~\cite{lego,dnc,epc,epc++}, and forces depth propagation from discriminative regions to textureless regions. 
However, such propagation is with limited range and tends to cause over-smooth results around boundaries.

Considering the basic limitation is from representation, feature-metric loss is proposed to use learned feature representation for each pixel, which is explicitly constrained to be discriminative even in textureless regions. For learning feature representation, a single view reconstruction pathway is added as an auto-encoder network. To ensure loss landscapes defined on the learned feature representation having desired shapes, two additional regularizing losses are added to the auto-encoder loss, \ie, discriminative loss and convergent loss. The discriminative loss encourages feature differences across pixels modeled by first-order gradients, while the convergent loss ensures a wide convergence basin by penalizing feature gradients' variances across pixels.


In total, our network architecture contains three sub-networks, \ie, DepthNet and PoseNet for cross-view reconstruction, and FeatureNet for single-view reconstruction, where features generated by FeatureNet are used to define feature-metric loss for DepthNet and PoseNet.

In experiment, feature-metric loss outperforms widely used first-order and second-order smoothness losses, and improves state-of-the-art depth estimation from 0.885 to 0.925 measured by $\delta_1$ on KITTI dataset. 
In addition, our method generates better egomotion estimation and results in more accurate visual odometry.

In general, our contributions are summarized as three-fold:
\begin{itemize}
\item Feature-metric loss is proposed for self-supervised depth and egomotion estimation.
\item FeatureNet is proposed for feature representation learning for depth and egomotion estimation. 
\item State-of-the-art performances on depth and egomotion estimation are achieved on KITTI dataset.
\end{itemize}
\section{Related Work}
\label{rw}

In this section, we review related works of self-supervised learning for two tasks, \ie, monocular depth and egomotion estimation, as well as visual representation learning.


\noindent
\textbf{Monocular depth and egomotion estimation:} 
SfMLearner is a pioneering work~\cite{sfmlearner} for this task, where geometry estimation from DepthNet and PoseNet is supervised by photometric loss. To tackle moving objects that break the assumption of static scenes, optical flow is estimated to compensate these moving pixels~\cite{GeoNet,epc,epc++,dfnet}, segmentation masks provided by pre-trained segmentation models are also to handle potential moving objects separately~\cite{struct2depth,signet,learnk}.

More geometric priors are also used to strengthen the self-supervised learning. Depth-normal consistency loss is proposed as as extra constraint~\cite{lego,dnc}. 3D consistency between point clouds backprojected from adjacent views is considered in~\cite{vid2deep,glnet,sc-sfmlearner}. In addition, binocular videos are used for training to solve both scale ambiguity and scene dynamics~\cite{undeepvo,monodepth2,epc,epc++}, where only inference can be carried on monocular video.

In contrast to all above methods where focuses are on the geometry parts of the task, deep feature reconstruction~\cite{dfr} proposed to use deep features from pre-trained models to define reconstruction loss. Our method shares the same spirit, but takes a step further to explicitly learn deep features from the geometry problem under the same self-supervised learning framework.

\noindent
\textbf{Visual representation learning:} 
It is of great interest of self-supervised visual representation learning for downstream tasks. Without explicitly provided labels, the losses are defined by manipulating the data itself in different ways, which could be reconstructing input data~\cite{stacked,denoise,afl,avb}, predicting spatial transformations~\cite{sp1,sp2,sp3,sp4}, coloring grayscale input images~\cite{colorization1,colorization2,colorization3,colorization4} etc. Our work belongs to reconstruct the input through an auto-encoder network. Different from previous works mainly aiming for learning better features for recognition tasks, our method is designed to learn better features for the geometry task.

\section{Method}
\label{method}
In this section, we firstly introduce geometry models with required notations, then define two reconstruction losses, one for depth and ego-motion learning, the other for feature representation learning.
Finally, we present our overall pipeline and implementation details about loss settings and network architectures.

\subsection{Geometry models}\label{sec41}
\textbf{Camera model and depth.}
The camera operator $\pi: \RR^3 \rightarrow \RR^2$ projects a 3D point $P=(X,Y,Z)$ to a 2D pixel $p=(u,v)$ by:
\begin{equation}\label{31}
    \pi(P) = (f_x \frac{X}{Z}+c_x, f_y \frac{Y}{Z} + c_y)
\end{equation}
where $(f_x, f_y, c_x, c_y)$ are the camera intrinsic parameters. Similarly, a pixel $p$ is projected to a 3D point $P$ given its depth $D(p)$, i.e., backprojection $\pi^{-1}: \RR^2 \times \RR \rightarrow \RR^3$:
\begin{equation}\label{32}
    \pi^{-1}\big(p, D(p)\big) = D(p)\Big(\frac{x-c_x}{f_x}, \frac{y-c_y}{f_y},1\Big)^\top
\end{equation}

\textbf{Ego-motion.} Ego-motion is modeled by transformation $G\in \SE(3)$, together with $\pi$ and $\pi^{-1}$, we can define a projective warping function $\omega: \RR^2 \times \RR \times \SE(3) \rightarrow \RR^2$, which maps a pixel $p$ in one frame to the other frame transformed by $G$:
\begin{equation}\label{warp}
     \widehat{p}=\omega\big(p,D(p),G\big)=\pi\Big(G\cdot \pi^{-1}\big(p,D(p)\big)\Big)
\end{equation}

\subsection{Cross-view reconstruction}
With the above geometry models, target frame $I_t$ can be reconstructed from source frame $I_s$ via,
\begin{equation}\label{sample1}
    \widehat{I}_{s \rightarrow t}(p) = I_s(\widehat{p})
\end{equation}
where $\widehat{p}$ is defined in Eq.~\ref{warp} and depends on both depth and ego-motion. 
$I_t(p)$ and $I_s(\widehat{p})$ should be similar given a set of assumptions, including both depth and ego-motion are correct; the corresponding 3D point is static with Lambertian reflectance and not occluded in both views. 
Then, a multi-view reconstruction loss can be defined for learning depth and motion, i.e.,
\begin{equation}\label{sample2}
    \LL_{s \rightarrow t} = \sum_p \ell\big(I_s(\widehat{p}), I_t(p)\big),
\end{equation}
where $\ell(,)$ is the per-pixel loss which measures the photometric difference, i.e, photometric loss.

Though the loss works, it is fundamentally problematic since correct depth and pose is sufficient but not necessary for small photometric error, e.g., pixels in a textureless with the same photometric values can have small photometric losses even the depth and pose are wrongly estimated.
The problem can be formally analysed from the optimization perspective by deriving the gradients with respect to both depth $D(p)$ and egomotion $G$, 

\begin{equation}\label{derivative1}
    \frac{\partial \LL_{s \rightarrow t}}{\partial D(p)} = \frac{\partial \ell\big(I_s(\widehat{p}), I_t(p)\big)}{\partial I_s(\widehat{p})} \cdot \frac{\partial I_s(\widehat{p})}{\partial \widehat{p}} \cdot \frac{\partial \widehat{p}}{\partial D(p)},
\end{equation}

\begin{equation}\label{derivative2}
    \frac{\partial \LL_{s \rightarrow t}}{\partial G} = \sum_p \frac{\partial \ell\big(I_s(\widehat{p}), I_t(p)\big)}{\partial I_s(\widehat{p})} \cdot \frac{\partial I_s(\widehat{p})}{\partial \widehat{p}} \cdot \frac{\partial \widehat{p}}{\partial G},
\end{equation}
where both gradients depend on the image gradient $\frac{\partial I_s(\widehat{p})}{\partial \widehat{p}}$. For textureless region, the image gradients are close to zero which further causes zero gradients for Eq.~\ref{derivative1} and contributes zero to Eq.~\ref{derivative2} for egomotion estimation. In addition, locally non-smooth gradient directions are also challenging convergence due to inconsistent update directions towards minima. 

Therefore, we propose to learn feature representation $\phi_s(p)$ with better gradient $\frac{\partial \phi_s(\widehat{p})}{\partial \widehat{p}}$ to overcome the above problems, and generalizes photometric loss to feature-metric loss accordingly,
\begin{equation}\label{feature_metric}
    \LL_{s \rightarrow t} = \sum_p \ell\big(\phi_s(\widehat{p}), \phi_t(p)\big).
\end{equation}

\subsection{Single-view reconstruction}
The feature representation $\phi(p)$ is also learned in self-supervised manner with single-view reconstruction through an auto-encoder network. The auto-encoder network contains an encoder for deep feature extractions from an image and an decoder to reconstruct the input image based on the deep features. The deep features are learned to encode large patterns in an image where redundancies and noises are removed. To ensure the learned representation with good properties for optimizing Eq.~\ref{feature_metric}, we add two extra regularizers $\LL_{dis}$ and $\LL_{cvt}$ to the image reconstruction loss $\LL_{rec}$, i.e.,
\begin{equation}\label{t}
    \LL_{s}=\LL_{rec}+\alpha \LL_{dis}+\beta \LL_{cvt}
\end{equation}
 where $\alpha$ and $\beta$ are set to 1e-3 via cross validation. These three loss terms are described in detail below.

For simplicity, we denote first-order derivative and second-order derivative with respect to image coordinates by $\nabla^1$ and $\nabla^2$, which equals $\partial_x+\partial_y$ and $\partial_{xx}+2\partial_{xy}+\partial_{yy}$ respectively.

\textbf{Image reconstruction loss}
Image reconstruction loss $\LL_{rec}$ is the standard loss function for an auto-encoder network, which requires the encoded features can be used to reconstruct its input, i.e.,
\begin{equation}\label{rec}
    \LL_{rec}= \sum_p |I(p) - I_{rec}(p)|_1
\end{equation}
where $I(p)$ is the input image, and $I_{rec}(p)$ is the image reconstructed from the auto-encoder network. 


\textbf{Discriminative loss} $\LL_{dis}$ is defined to ensure the learned features have gradients $\frac{\partial \phi(\widehat{p})}{\partial \widehat{p}}$ by explicitly encouraging large gradient, i.e.,
\begin{equation}\label{dis1}
    \LL_{dis}=-\sum_p |\nabla^1 \phi(p)|_1
\end{equation}
Furthermore, image gradients are used to emphasize low-texture regions,  
\begin{equation}\label{dis2}
    \LL_{dis}=-\sum_p e^{-|\nabla^1 I(p)|_1} |\nabla^1 \phi(p)|_1
\end{equation}
where low-texture regions receive large weights. 


\textbf{Convergent loss} $\LL_{cvt}$ is defined to encourage smoothness of feature gradients, which ensures consistent gradients during optimization and large convergence radii accordingly. The loss is defined to penalize the second-order gradients, i.e.,
\begin{equation}\label{cvt1}
    \LL_{cvt}=\sum_p |\nabla^2 \phi(p)|_1
\end{equation}





\begin{figure*}[!tp]
\centering
\includegraphics[width=12cm]{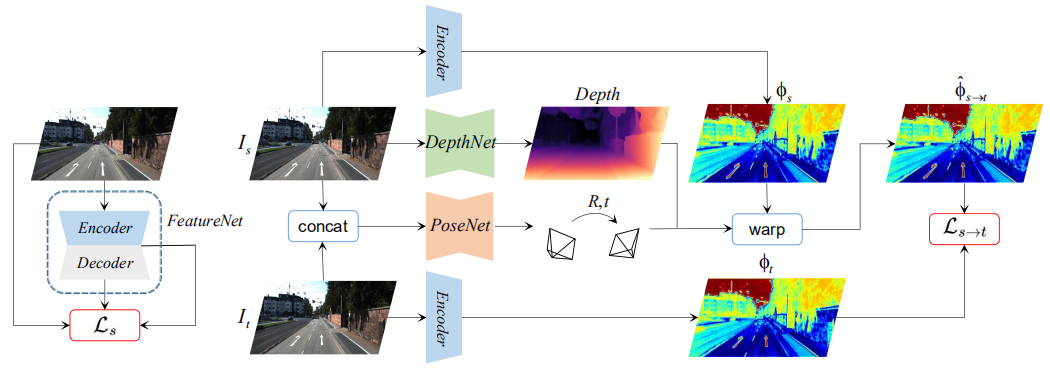}
\caption{
An illustration of the overall framework, which contains DepthNet, PoseNet and FeatureNet for depth map prediction, egomotion prediction and feature learning respectively.
FeatureNet uses $\LL_{s}$ to learn require visual representation, the encoder from FeatureNet is used to extract features for cross-view reconstruction loss $\LL_{s \rightarrow t}$.
}
\label{fig3}
\end{figure*}

\subsection{Overall pipeline}\label{sec34}
Single-view reconstruction and cross-view reconstruction are unified to form the final framework as illustrated in Fig.~\ref{fig3}. DepthNet is a monodepth estimator which takes the target frame as input and outputs a depth map. PoseNet is an egomotion estimator, which takes two frames from both source and target view and outputs the relative pose between them. DepthNet and PoseNet provide the geometry information to establish point-to-point correspondences for cross-view reconstruction. FeatureNet is for feature representation learning, which follows the auto-encoder architecture and supervised by single-view reconstruction loss. Features from FeatureNet are used to define the cross-view reconstruction loss.   

Therefore, the total loss for the whole architecture contains two parts, where $\LL_s$ constrains the quality of learned features through single-view reconstruction, whilst $\LL_{s \rightarrow t}$ penalizes the discrepancy from cross-view reconstruction, i.e., 
\begin{equation}
    \LL_{total} = \LL_s + \LL_{s \rightarrow t}
\end{equation}

Toward better performance, the proposed feature-metric loss is combined with used photometric loss, i.e.,
\begin{equation}\label{fmph}
\begin{aligned}
    \LL_{s \rightarrow t} = &\sum_p  \LL_{fm}\big(\phi_s(\widehat{p}),\phi_t(p)\big)\\
    +&\sum_p \LL_{ph}(I_s(\widehat{p}),I_t(p))
\end{aligned}
\end{equation}
where $\LL_{fm}$ and $\LL_{ph}$ are the feature-metric loss and photometric loss respectively. Specifically, feature-metric loss is defined by
\begin{equation}
\LL_{fm} = |\phi_s(\widehat{p})- \phi_t(p)|_1,
\end{equation}
and photometric loss is defined following~\cite{monodepth} using a combination of $L_1$ and SSIM
losses, i.e.,
\begin{equation}\label{ph}
\begin{aligned}
\LL_{ph} =0.15 &\sum_p |I_s(\widehat{p})-I_t(p)|_1+\\
0.85 &\frac{1-\text{SSIM}(I_s(\widehat{p}),I_t(p))}{2}
\end{aligned}
\end{equation}

Furthermore, we resolve the occlusion problem following the practices in~\cite{monodepth2,ddvo,noise,dfr}, where two source views are used to define the cross-view reconstruction loss,
\begin{equation}\label{occ}
\begin{aligned}
    \LL_{s \rightarrow t}' = &\sum_p \underset{s \in V}{\min} \; \LL_{s \rightarrow t}\big(\phi_s(\widehat{p}),\phi_t(p)\big)
\end{aligned}
\end{equation}
Where $V$ is a set composed of source frames.
When trained on the monocular videos, $V$ contains the previous and posterior source frames of current target frame; when trained on the calibrated binocular videos, an extra frame of opposite stereo pair is added.

\begin{figure*}[!tp]
\centering
\includegraphics[width=12cm]{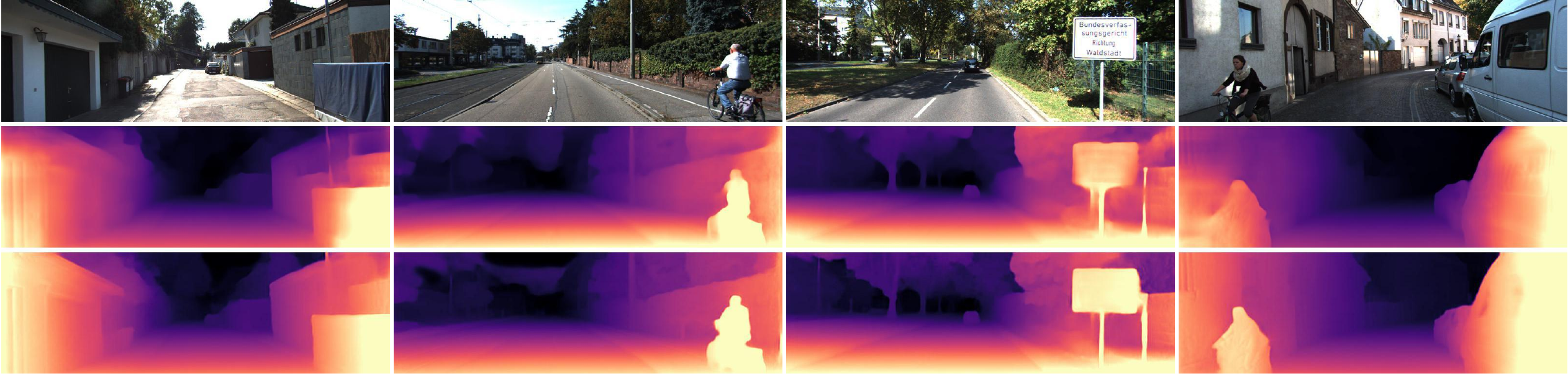}
\caption{
Qualitative comparison between Monodepth2 \cite{monodepth2} (second row) and our method (last row).
It can be seen that we achieve better performance on the low-texture regions like walls and billboards, and finer details are present like silhouette of humans and poles.
}
\label{qualitative}
\end{figure*}

\subsection{Implementation details}\label{sec45}
For FeatureNet, ResNet-50~\cite{resnet} with fully-connected layer removed is used as the encoder, where deepest feature map goes through 5 downsampling stages and reduces to 1/32 resolution of input image, the decoder contains five $3\times 3$ convolutional layers and each followed by a bilinear upsampling layer. Multi-scale feature maps from convolutional layers of the decoder are used to generate multi-scale reconstructed images, where feature map of each scale further goes through a $3\times 3$ convolution with sigmoid function for image reconstruction. The largest feature map with 64 channels from encoder is regularized by $\LL_{dis}$ and $\LL_{cvt}$ and will be used for feature-metric loss.

DepthNet also adopts an encoder-decoder structure, where ResNet-50 without fully-connected layer is used as encoder and multi-scale feature maps are outputted. The decoder for depth is implemented in a cascaded refinement manner, which decodes depth maps in a top-down pathway. Specifically, multiple-scale features from encoder are used to predict maps of corresponding sizes via a $3\times 3$ convolution followed by sigmoid, and these maps are refined in a coarse-to-fine manner towards the final depth map.  Both FeatureNet and DepthNet take image size of $320 \times 1024$ as inputs.

The PoseNet is a pose estimator with a structure of ResNet-18 \cite{resnet}, which is modified to receive a concatenated image pair and predicts a relative pose therein. Here axis angle is chosen to represent the 3D rotation.
The input resolution is $192 \times 640$. Comparing with both FeatureNet and DepthNet, PoseNet uses lower image resolution and more light-weight backbone, which observes this has no obvious influence to pose accuracy, but significantly save both memory and computation.

We adopt the setting in~\cite{monodepth2} for data preprocessing.
Our models are implemented on PyTorch \cite{pytorch} with distributed computing, and trained for 40 epochs using Adam~\cite{adam} optimizer, with a batch size of 2, on the 8 GTX 1080Ti GPUs.
The learning rate is gradually warmed up to $1e^{-4}$ in 3 steps, where each step increases learning rate by $1e^{-4}/3$ in 500 iterations. After warmping, learning rate $1e^{-4}$ is used for the first 20 epochs and halved twices at 20th and 30th epoch.
As for online refinement technique we used during testing, we follow the practice proposed by \cite{glnet,struct2depth}. 
We keep the model training while performing inference.
The batch size is set to 1. 
Each batch consists of the test image and its two adjacent frames. 
Online refinement is performed for 20 iterations on one test sample with the same setting introduced before. 
No data augmentation is used in the inference phase.

\section{Experiments}
\label{exp}

In this section we show extensive experiments for evaluating the performance of our approach. We make a fair comparison on KITTI 2015 dataset~\cite{kitti} with prior art on both single view depth and visual odometry estimation tasks. 
And detailed ablation studies of our approach are done to show the effectiveness of the \textbf{feature-metric loss}.
    
KITTI 2015 dataset contains videos in 200 street scenes captured by RGB cameras, with sparse depth ground truths captured by Velodyne laser scanner.
We follow \cite{sfmlearner} to remove static frames as pre-processing step.
We use the Eigen split of \cite{Eigen} to divide KITTI raw data, and resulting in 39,810 monocular triplets for training, 4,424 for validation and 697 for testing.

For depth evaluation, we test our depth model on divided 697 KITTI testing data. 
For odometry evaluation, we test our system to the official KITTI odometry split which containing 11 driving sequences with ground truth odometry obtained through the IMU and GPS readings. 
Following previous works \cite{dfr,sc-sfmlearner,sfmlearner}, we train our model on the sequence 00-08 and use the sequence 09-10 for testing.

\subsection{Depth evaluation.}
\begin{table}[!tp]
\begin{center}
\begin{tabular}{ll}
\hline
$\textbf{Abs Rel}:\frac{1}{|D|}\sum_{d \in D}|d^*-d|/d^*$ 
&$\textbf{RMSE}:\sqrt{\frac{1}{|D|}\sum_{d \in D}||d^*-d||^2}$\\

$\textbf{Sq Rel}:\frac{1}{|D|}\sum_{d \in D}||d^*-d||^2/d^*$
&$\textbf{RMSE log}:\sqrt{\frac{1}{|D|}\sum_{d \in D}||logd^*-logd||^2}$\\

\multicolumn{2}{l}{$\mathbf{\delta}_\mathbf{t}:\frac{1}{|D|}|\{d \in D| \: max(\frac{d^*}{d},\frac{d}{d^*}) \: < 1.25^t\}|\times 100\%$} 
\\
\hline
\end{tabular}
\caption{Performance metrics for depth evaluation.
$d$ and $d^*$ respectively denotes predicted and ground truth depth, $D$ presents a set of all the predicted depth values of an image, $|.|$ returns the number of the elements in the input set.
}
\label{tab1}
\end{center}
\end{table}

\textbf{Performance metrics.}
Standard metrics are used for depth evaluation, as shown in Tab.~\ref{tab1}. 
During evaluation, depth is capped to 80m.
For the methods trained on monocular videos, the depth is defined up to scale factor \cite{sfmlearner}, which is computed by
\begin{equation}\label{scale}
scale = median(D_{gt})/median(D_{pred})
\end{equation}
For evaluation, those predicted depth maps are multiplied by computed $scale$ to match the median with the ground truth, this step is called \textbf{median scaling}.

\begin{table}[!t]
\begin{center}
\resizebox{0.9\textwidth}{!}{
\begin{tabular}{l|l|cccc|ccc}
\hline
\multirow{2}*{Method} &\multirow{2}*{Train} &\multicolumn{4}{c|}{The lower the better} &\multicolumn{3}{c}{The higher the better}\\
~ &~ &Abs Rel &Sq Rel &RMSE &RMSE log &$\delta_1$ &$\delta_2$ &$\delta_3$\\
\hline
\hline
SfMLearner~\cite{sfmlearner}       &M &0.208 &1.768 &6.958 &0.283 &0.678 &0.885 & 0.957\\
DNC~\cite{dnc}                     &M &0.182 &1.481 &6.501 &0.267 &0.725 &0.906 &0.963\\
Vid2Depth~\cite{vid2deep}          &M &0.163 &1.240 &6.220 &0.250 &0.762 &0.916 &0.968\\
LEGO~\cite{lego}                   &M &0.162 &1.352 &6.276 &0.252 &0.783 &0.921 &0.969\\
GeoNet~\cite{GeoNet}               &M &0.155 &1.296 &5.857 &0.233 &0.793 &0.931 &0.973\\
DF-Net~\cite{dfnet}                &M &0.150 &1.124 &5.507 &0.223 &0.806 &0.933 &0.973\\
DDVO~\cite{ddvo}                   &M &0.151 &1.257 &5.583 &0.228 &0.810 &0.936 &0.974\\
EPC++~\cite{epc++}                 &M &0.141 &1.029 &5.350 &0.216 &0.816 &0.941 &0.976\\
Struct2Depth~\cite{struct2depth}   &M &0.141 &1.036 &5.291 &0.215 &0.816 &0.945 &0.979\\
SIGNet~\cite{signet}               &M &0.133 &0.905 &5.181 &0.208 &0.825 &0.947 &0.981\\
CC~\cite{cc}                       &M &0.140 &1.070 &5.326 &0.217 &0.826 &0.941 &0.975\\
LearnK~\cite{learnk}               &M &0.128 &0.959 &5.230  &0.212 &0.845 &0.947 &0.976\\
DualNet~\cite{dualnet} &M 
&0.121 &\underline{0.837} &4.945 &0.197 &0.853 &0.955 &\underline{0.982}\\
SuperDepth~\cite{superdepth}       &M &0.116 &1.055 &-     &0.209 &0.853 &0.948 &0.977\\
Monodepth2~\cite{monodepth2}       &M &\underline{0.115} &0.882 &\underline{4.701} &\underline{0.190} &\underline{0.879} &\underline{0.961} &\underline{0.982}\\ 
\hline
Ours                               &M &\textbf{0.104} &\textbf{0.729} &\textbf{4.481} &\textbf{0.179} &\textbf{0.893} &\textbf{0.965} &\textbf{0.984}\\
\hline
\hline
Struct2Depth~\cite{struct2depth}   &M${^*}$ &0.109 &0.825 &4.750 &0.187 &0.874 &\underline{0.958} &\textbf{0.983}\\
GLNet~\cite{glnet} &M$^{*}$ &\underline{0.099} &\underline{0.796} &\underline{4.743} &\underline{0.186} &\underline{0.884} &0.955 &0.979\\
\hline
Ours                           &M$^{*}$ &\textbf{0.088} &\textbf{0.712} &\textbf{4.137} &\textbf{0.169} &\textbf{0.915} &\textbf{0.965} &\underline{0.982}\\
\hline
\hline
Dorn~\cite{dorn} &Sup &0.099 &0.593 &3.714 &0.161 &0.897 &0.966 &0.986\\
BTS~\cite{bts} &Sup &0.091 &0.555 &4.033 &0.174 &0.904 &0.967 &0.984\\
\hline
\hline
MonoDepth~\cite{monodepth}         &S &0.133 &1.142 &5.533 &0.230 &0.830 &0.936 &0.970\\
MonoDispNet~\cite{monodispnet}     &S &0.126 &0.832 &\textbf{4.172} &0.217 &0.840 &0.941 &0.973\\
MonoResMatch~\cite{monoresmatch}   &S &0.111 &0.867 &4.714 &0.199 &0.864 &0.954 &\underline{0.979}\\
MonoDepth2~\cite{monodepth2}       &S &0.107 &0.849 &4.764 &0.201 &0.874 &0.953 &0.977\\
RefineDistill~\cite{refinedistill} &S &\textbf{0.098} &0.831 &4.656 &0.202 &0.882 &0.948 &0.973\\
UnDeepVO~\cite{undeepvo}           &MS &0.183 &1.730 &6.570 &0.268 &- &- &-\\
DFR~\cite{dfr}                     &MS &0.135 &1.132 &5.585 &0.229 &0.820 &0.933 &0.971\\
EPC++~\cite{epc++}                 &MS &0.128 &0.935 &5.011 &0.209 &0.831 &0.945 &\underline{0.979}\\
MonoDepth2~\cite{monodepth2}       &MS &0.106 &0.818 &4.750 &0.196 &0.874 &0.957 &\underline{0.979}\\
DepthHint~\cite{depthhint}         &MS$^\dagger$ &0.100 &\underline{0.728} &4.469 &\underline{0.185} &\underline{0.885} &\underline{0.962} &\textbf{0.982}\\
\hline
Ours                               &MS &\underline{0.099} &\textbf{0.697} &\underline{4.427} &\textbf{0.184} &\textbf{0.889} &\textbf{0.963} &\textbf{0.982}\\
\hline
\hline
Ours                  &MS$^{*}$ &0.079 &0.666 &3.922 &0.163 &0.925 &0.970 &0.984\\
\hline
\end{tabular}
}
\caption{
Comparison of performances are reported on the KITTI dataset. 
Best results are in bold, second best are underlined.
M: trained on monocular videos.
S: trained on stereo pairs.
MS: trained on calibrated binocular videos.
Sup: trained on labelled single images.
$*$: using the online refinement technique \cite{struct2depth}, which advocated keeping the model training while performing inference.
$\dagger$: using post processing steps.
}
\label{tab2}
\end{center}
\end{table}
\vspace{2pt}
\textbf{Comparison with state-of-the-art.}
Tab.~\ref{tab2} shows performances of current state-of-the-art approaches for monocular depth estimation. 
They are trained on different kinds of data --- monocular videos (M), stereo pairs (S), binocular videos (MS) and labelled single images (Sup), while all of them are tested with single image as input.

We achieve the best performance compared to all self-supervised methods, no matter which training data is used. 
Our method achieves more significant improvement in the performance metric Sq Rel. According to Tab. \ref{tab1}, this metric penalizes more on large errors in short range, where more textureless regions exist due near objects are large in images and our method handles well.
The closest results in self-supervised methods are from DepthHint~\cite{depthhint}, which uses the same input size but adds an extra post processing step.
It utilizes a traditional stereo matching method --- SGM~\cite{sgm} to provide extra supervisory signals for training, since SGM is less likely to be trapped by local minimums.
However, in its settings, the object function of SGM is still photometric loss, the drawbacks of photometric loss are still inevitable.
In contrast, proposed feature-metric loss will largely avoid the interference of local minimums.

Moreover, compared with state-of-the-art \textbf{supervised} methods \cite{dorn,bts}, which achieve top performances on the KITTI depth prediction competition, our model with online refinement technique even exceeds in many metrics. 
Our advantage over supervised methods is that the gap between the distributions of training and testing data does exist, we can make full use of online refinement technique.
What is more, as shown in Sec. \ref{sec43}, the introduction of feature-metric loss can obtain more performance gain from online refinement technique.

Fig.~\ref{qualitative} shows the qualitative results. 
Compared with state-of-the-art method MonoDepth2~\cite{monodepth2}, we achieve better performance on low-texture regions and finer details, e.g., walls, billboards, silhouette of humans and poles.

However, MonoDepth2 is built on the photometric loss, which is easily trapped by local minimums especially on low-texture regions like walls and billboards.
In contrast, the introduction of feature-metric loss leads the network into jumping out of local minimums, since our features are designed to form a desirable loss for easier optimization.

\begin{table}[!t]
\begin{center}
\begin{tabular}{l|cc|cc}
\hline
\multirow{2}*{Method} &\multicolumn{2}{c|}{Seq. 09} &\multicolumn{2}{c}{Seq. 10}\\
~ &$t_{err}$ &$r_{err}$ &$t_{err}$ &$r_{err}$\\
\hline
ORB-SLAM \cite{orbslam} &15.30 &0.26 &3.68 &0.48 \\
\hline
SfMLearner \cite{sfmlearner} &17.84 & 6.78 &37.91 &17.78 \\
DFR \cite{dfr} &11.93 &3.91 &12.45 &3.46 \\
MonoDepth2 \cite{monodepth2}  &10.85 &2.86 &11.60 &5.72 \\
NeuralBundler \cite{neuralbundler} &\textbf{8.10} &2.81 &12.90 &\textbf{3.17} \\
SC-SfMlearner \cite{sc-sfmlearner}  &8.24 &2.19 &10.70 &4.58 \\
\hline
Ours &8.75 &\textbf{2.11} &\textbf{10.67} &4.91\\
\hline
\end{tabular}
\caption{Comparison of performances are reported on the KITTI odometry dataset \cite{kitti}. 
Best results are in bold.}
\label{tab4}
\end{center}
\end{table}

\subsection{Odometry evaluation}\label{sec53}
\textbf{Performance metric.} 
Average translational root mean square error drift ($t_{err}$) and average rotational root mean square error drift ($r_{err}$) on length of 100m - 800m are adopted for evaluation.
For the methods who suffer from scale ambiguity, one global scale that best align the whole sequence is used. 

\textbf{Comparison with state-of-the-art.}
As shown in Tab. \ref{tab4}, we report the performance of ORB-SLAM\cite{orbslam} as a reference and compare with recent deep methods.
our method gets top performances in two metrics and comparable performance in the rest metrics compared to other deep learning methods. 
When compared to traditional SLAM method \cite{orbslam}, our translation performance is comparable, while in the rotation estimation we still fall short like other deep learning methods.
We believe that it is because the bundle adjustment of the traditional SLAM method can optimize subtler rotation errors along a long sequence which can't be observed in a small sequence used by current deep learning based methods.  
Moreover current reconstruction process may be not sensible to variation of rotation \cite{bian2020depth}.


\begin{table*}[!tp]
\begin{center}
\resizebox{0.8\textwidth}{!}{
\begin{tabular}{l|c|cccc|ccc}
\multirow{2}*{Method} &\multirow{2}*{OR} &\multicolumn{4}{c|}{The lower the better} &\multicolumn{3}{c}{The higher the better}\\
~ &~ &Abs Rel &Sq Rel &RMSE &RMSE log &$\delta_1$ &$\delta_2$ &$\delta_3$\\
\hline
$\LL_{ph}+\LL_{ds}^1$ &$\times$
&0.105 &0.748 &4.835 &0.191 &0.878 &0.956 &0.979\\
$\LL_{ph}+\LL_{ds}^{2}$ &$\times$
&0.103 &0.740 &4.754 &0.187 &0.881 &0.959 &0.981\\
$\LL_{ph}+\LL_{ds}^{1}+\LL_{ds}^{2}$ &$\times$
&0.103 &0.735 &4.554 &0.187 &0.883 &0.961 &0.981\\
$\LL_{ph}+\LL_{ds}^{1}+\LL_{ds}^{2}$ &$\checkmark$
&0.088 &0.712 &4.237 &0.175 &0.905 &0.965 &0.982\\
\hline
$\LL_{ph}+\LL_{fm}$ &$\times$
&0.099 &0.697 &4.427 &0.184 &0.889 &0.963 &0.982\\
$\LL_{ph}+\LL_{fm}$ &$\checkmark$
&\textbf{0.079} &\textbf{0.666} &\textbf{3.922} &\textbf{0.163} &\textbf{0.925} &\textbf{0.970} &\textbf{0.984}\\
\end{tabular}
}
\caption*{
(a)
\textbf{Different loss combinations in} $\LL_{s \rightarrow t}$ (Eq. \ref{feature_metric}), the term 'OR' denotes whether the online refinement \cite{struct2depth} is used.
}
\resizebox{0.98\textwidth}{!}{
\begin{tabular}{l|cccc|ccc|cc|cc}
\multirow{2}*{Loss} &\multicolumn{4}{c|}{The lower the better} &\multicolumn{3}{c|}{The higher the better} &\multicolumn{2}{c|}{Seq. 09} &\multicolumn{2}{c}{Seq. 10}\\
~ &Abs Rel &Sq Rel &RMSE &RMSE log &$\delta_1$ &$\delta_2$ &$\delta_3$ &$t_{err}$ &$r_{err}$ &$t_{err}$ &$r_{err}$\\
\hline
$\LL_{rec}$
&0.105 &0.739 &4.585 &0.191 &0.883 &0.961 &\textbf{0.982} &4.30 &1.18 &8.50 &4.06\\
$\LL_{rec}+\LL_{dis}$
&0.103 &0.723 &4.535 &0.187 &0.884 &0.961 &\textbf{0.982}
&4.10 &1.07 &8.03 &3.94\\
$\LL_{rec}+\LL_{cvt}$
&0.100 &0.721 &4.474
&0.187
&0.885
&0.962
&\textbf{0.982}
&3.29
&1.16
&5.91
&3.48\\
$\LL_{rec}+\LL_{dis}+\LL_{cvt}$
&\textbf{0.099} &\textbf{0.697} &\textbf{4.427} &\textbf{0.184} &\textbf{0.889} &\textbf{0.963} &\textbf{0.982} &\textbf{3.07} &\textbf{0.89} &\textbf{3.83} &\textbf{1.78}\\
\end{tabular}
}
\caption*{
(b)
\textbf{Different loss combinations in $\LL_{s}$} (Eq. \ref{t}).
}
\caption{
The ablation study of different loss settings of our work.
}
\label{tab3}
\end{center}
\end{table*}

\subsection{Ablation study}\label{sec43}
To get a better understanding of the contribution of proposed losses---feature-metric loss, discriminative loss and convergent loss---to the overall performance, we perform an ablation study in Tab. \ref{tab3}.

\textbf{The losses for cross-view reconstruction.}
In Tab. \ref{tab3}a, different components of $\LL_{s \rightarrow t}$ have been tried.
The smoothness losses which are widely used are used as baselines:
\begin{equation}
    \LL_{ds}^i=
    \sum_p e^{-|\nabla^i I(p)|_1} |\nabla^i \widehat{D}(p)|_1
\end{equation}
where $\widehat{D}(p) = D(p)/\bar{D}$, this operation is the mean normalization technique advocated by \cite{ddvo}.
$i$ denotes the order of the derivatives.
These smoothness losses are used as baselines to verify the effectiveness of the feature-metric loss. 

Compared with smoothness losses, feature-metric loss leads to much better effect.
We can see that a biggest performance boost is gained by introducing the feature-metric loss.
As we discussed before, the propagation range of smoothness losses is limited, in contrast, the feature-metric loss enable a long-range propagation, since it has a large convergence radius.
We also observe that when feature-metric loss can benefit more from the performance gain provided by online refinement than other loss combination.
Higher performance gain is attributed to better supervised signal provided by feature-metric loss during online refinement phase, where incorrect depth values can be appropriately penalized with larger losses based on more discriminative features.

\begin{figure}[!t]
\centering
\includegraphics[width=12cm]{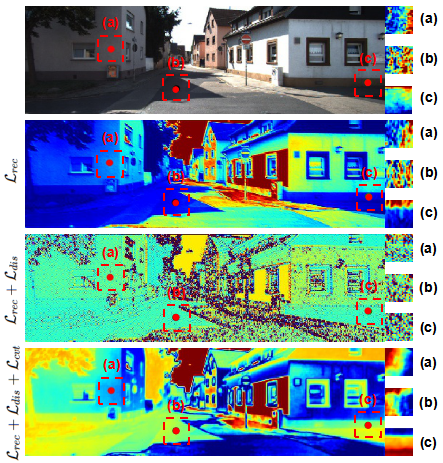}
\caption{
A visualization of a learned visual representation, which is achieved by selecting one principle channel through PCA decomposition, then showing the feature map as a heat map, hotter color indicates a higher feature value.
First row shows a typical image which is full of textureless regions like walls and shadows.
The visualization of corresponding feature maps is shown in second to fourth rows.
The feature maps are respectively learned with different loss combinations, which sequentially correspond with the settings in the first three rows in Tab. \ref{tab3}b.
In order to get a better understanding, we crop three typical textureless regions as shown in (a-c), cropped feature maps are visualized according to the dynamic range after cropping.
}
\label{feature}
\end{figure}

\textbf{The losses for single-view reconstruction.}
Tab.\ref{tab3}b shows that the model without any of our contributions performs the worst. 
When combined together, all our components lead to a significant improvement.

And as shown in right part of Tab. \ref{tab3}b, although small deviations are less obvious in some metrics of the depth evaluation, small errors will be magnified via accumulation and propagation during trajectory prediction, big differences are shown in the odometry evaluation.
Note that different from previous odometry evaluation, we directly applied the model trained on the kitti raw data to sequence 09-10 to get $t_{err}$ and $r_{err}$.

Merely using $\LL_{rec}$ gets similar performance as merely using photometric loss (the third row in Tab. \ref{tab3}a), since it plays a similar role as the photometric loss at textureless regions.
Results get better when equipped with $\LL_{dis}$, since discrimination at low-texture regions is improved.
Best performance is achieved when added $\LL_{cvt}$, which means discrimination is not enough, a correct optimization direction is also important.

\textbf{Visualization analysis.}
In order to see whether learned visual representations have promised properties, we visualize it in Fig. \ref{feature}.
The feature maps learned with different loss combinations: $\LL_{rec}$, $\LL_{rec}+\LL_{dis}$ and $\LL_{rec}+\LL_{dis}+\LL_{cvt}$ are sequentially shown from the second to the fourth row.
Although we require our feature to be discriminative, this effect is not sufficient to be shown in a large view, since the gap between the features of different sorts are much larger than that of spatially adjacent features.
Therefore, we cropped three typical textureless regions, and visualize them again according to the dynamic range after cropping. 

We can see that merely using $\LL_{rec}$ get small variations at textureless regions.
The close-ups of original images are similar to feature maps only trained with $\mathcal{L}_{rec}$, which verifies the proposed losses in improving feature representations.
The feature map learned with $\LL_{rec}+\LL_{dis}$ is not smooth and disordered, since $\LL_{dis}$ overemphasizes the discrepancy between adjacent features, the network degenerates to form a landscape of a zigzag shape.
This phenomenon can be approved by the results in the second row of Tab. \ref{tab3}b, which is only slightly higher than merely using $\LL_{rec}$.

A desired landscape for feature maps is a smooth slope, in this way, feature-metric loss will be able to form a basin-like landscape.
The feature map learned with all the proposed losses approximates this ideal landscape, from zoom-in views we can see a clear and smooth transition along a certain direction.
On this landscape, gradient descent approaches can move smoothly toward optimal solutions.
\section{Conclusion}
\label{con}
In this work, feature-metric loss is proposed for self-supervised learning of depth and egomotion, where feature representation is additionally learned with two extra regularizers to ensure convergence towards correct depth and pose. The whole framework is end-to-end trainable in self-supervised setting, and achieves state-of-the-art depth estimation which is even comparable to supervised learning methods. Furthermore, visual odometry based on estimated egomotion also significantly outperforms previous state-of-the-art methods.
\\
\noindent
\textbf{Acknowledgements} This research is supported by Beijing Science and Technology Project (No. Z181100008918018).


\clearpage
%
%
\bibliographystyle{splncs04}
\bibliography{reference}
\end{document}